\documentclass[runningheads]{llncs}
\usepackage{graphicx}
\graphicspath{{./images/}}
\DeclareGraphicsExtensions{.pdf,.png}

\usepackage{amsmath,amssymb} % define this before the line numbering.
\usepackage{color}

\usepackage[hidelinks]{hyperref}

\usepackage{multirow}
\usepackage{ctable} % nicer tables
\usepackage{booktabs} % nicer tables

\usepackage{algorithm}
\usepackage[noend]{algpseudocode}

\usepackage[caption=false]{subfig} % tables and figures side by side

\begin{document}
\pagestyle{headings}
\def\ECCV18SubNumber{2538}  % Insert your submission number here
\mainmatter

%TODO: choose better name
\title{A Dataset for Lane Instance Segmentation in Urban Environments} % Replace with your title
% Replace with your title

\titlerunning{A Dataset for Lane Instance Segmentation in Urban Environments}
% Replace with a meaningful short version of your title

\authorrunning{B. Roberts et al.}
% Replace with shorter version of the author list. If there are more authors than fits a line, please use A. Author et al.

\author{Brook Roberts \and Sebastian Kaltwang \and Sina Samangooei\and \\ Mark Pender-Bare \and Konstantinos Tertikas \and John Redford}

%Please write out author names in full in the paper, i.e. full given and family names. 
%If any authors have names that can be parsed into FirstName LastName in multiple ways, please include the correct parsing, in a comment to the volume editors:
%\index{Lastnames, Firstnames}
%(Do not uncomment it, because you may introduce extra index items if you do that, we will use scripts for introducing index entries...)

\institute{FiveAI Ltd., Cambridge CB2 1NS, U.K.\\
    \email{ \{brook,sebastian,sina,mark.pender-bare,konstantinos,john\}@five.ai}
}

\maketitle

\begin{abstract}
Autonomous vehicles require knowledge of the surrounding road layout, which can be predicted by state-of-the-art CNNs. 
This work addresses the current lack of data for determining lane instances, which are needed for various driving manoeuvres. 
The main issue is the time-consuming manual labelling process, typically applied per image. 
We notice that driving the car is itself a form of annotation.
Therefore, we propose a semi-automated method that allows for efficient labelling of image sequences by utilising an estimated road plane in 3D based on where the car has driven and projecting labels from this plane into all images of the sequence. 
The average labelling time per image is reduced to 5 seconds and only an inexpensive dash-cam is required for data capture. 
We are releasing a dataset of 24,000 images and additionally show experimental semantic segmentation and instance segmentation results.
\end{abstract}

\section{Introduction}
Autonomous vehicles have the potential to revolutionise urban transport. Mobility will be safer, always available, more reliable and provided at a lower cost. Yet we are still at the beginning of implementing fully autonomous systems, with many unsolved challenges remaining \cite{Janai2017}. 
One important problem is giving the autonomous system knowledge about surrounding space: a self-driving car needs to know the road layout around it in order to make informed driving decisions. 
In this work, we address the problem of detecting driving lane instances from a camera mounted on a vehicle. Separate, space-confined lane instance regions are needed to perform various challenging driving manoeuvres, including lane changing, overtaking and junction crossing.

Typical state-of-the-art CNN models need large amounts of labelled data to detect lane instances reliably (e.g. \cite{huval2015empirical,oliveira2016efficient,neven2018towards}).
However, few labelled datasets are publicly available, mainly due to the time consuming annotation process; it takes from several minutes up to more than one hour per image \cite{brostow2009semantic,cordts2016cityscapes,neuhold2017mapillary} to annotate images completely for semantic segmentation tasks.
In this work, we introduce a new video dataset for road segmentation, ego lane segmentation and lane instance segmentation in urban environments. We propose a semi-automated annotation process, that reduces the average time per image to the order of seconds. This speed-up is achieved by (1) noticing that driving the car is itself a form of annotation and that cars mostly travel along lanes, (2) propagating manual label adjustments from a single view to all images of the sequence and (3) accepting non-labelled parts in ambiguous situations.

% lane instance detection approaches
Previous lane detection work has focused on detecting the components of lane boundaries, and then applying clustering to identify the boundary as a whole \cite{mccall2006video,kim2008robust,gopalan2012learning,huval2015empirical}.
More recent methods use CNN based segmentation \cite{huval2015empirical,neven2018towards}, and RNNs \cite{li2017deep} for detecting lane boundaries.
However, visible lane boundaries can be interrupted by occlusion or worn markings, and by themselves are not associated with a specific lane instance.
Hence, we target lane instance labels in our dataset, which provide a consistent definition of the lane surface (from which lane boundaries can be derived).
Some work focuses on the road markings \cite{mathibela2015reading}, which are usually present at the border of lanes. However, additional steps are needed to determine the area per lane.
Much of the work has only been evaluated on proprietary datasets and only few public datasets are available \cite{hillel2014recent}.
\begin{figure}[htbp]
	\centering
	\includegraphics[width=0.45\linewidth]{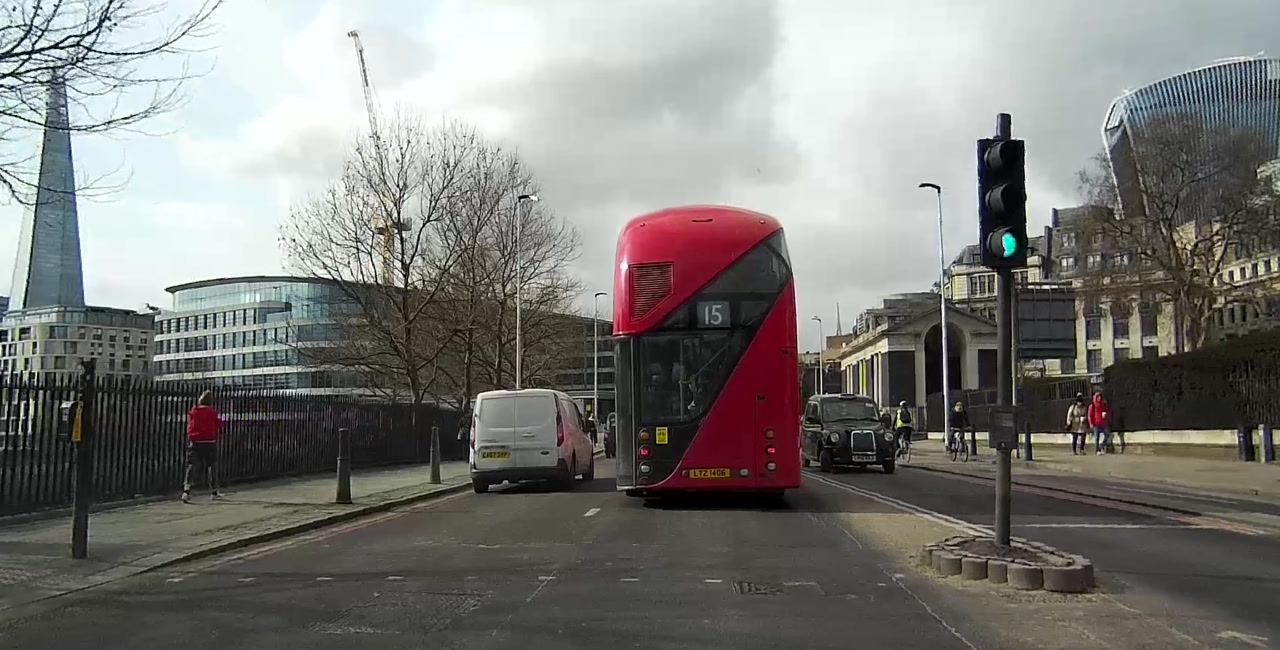}
	\includegraphics[width=0.45\linewidth]{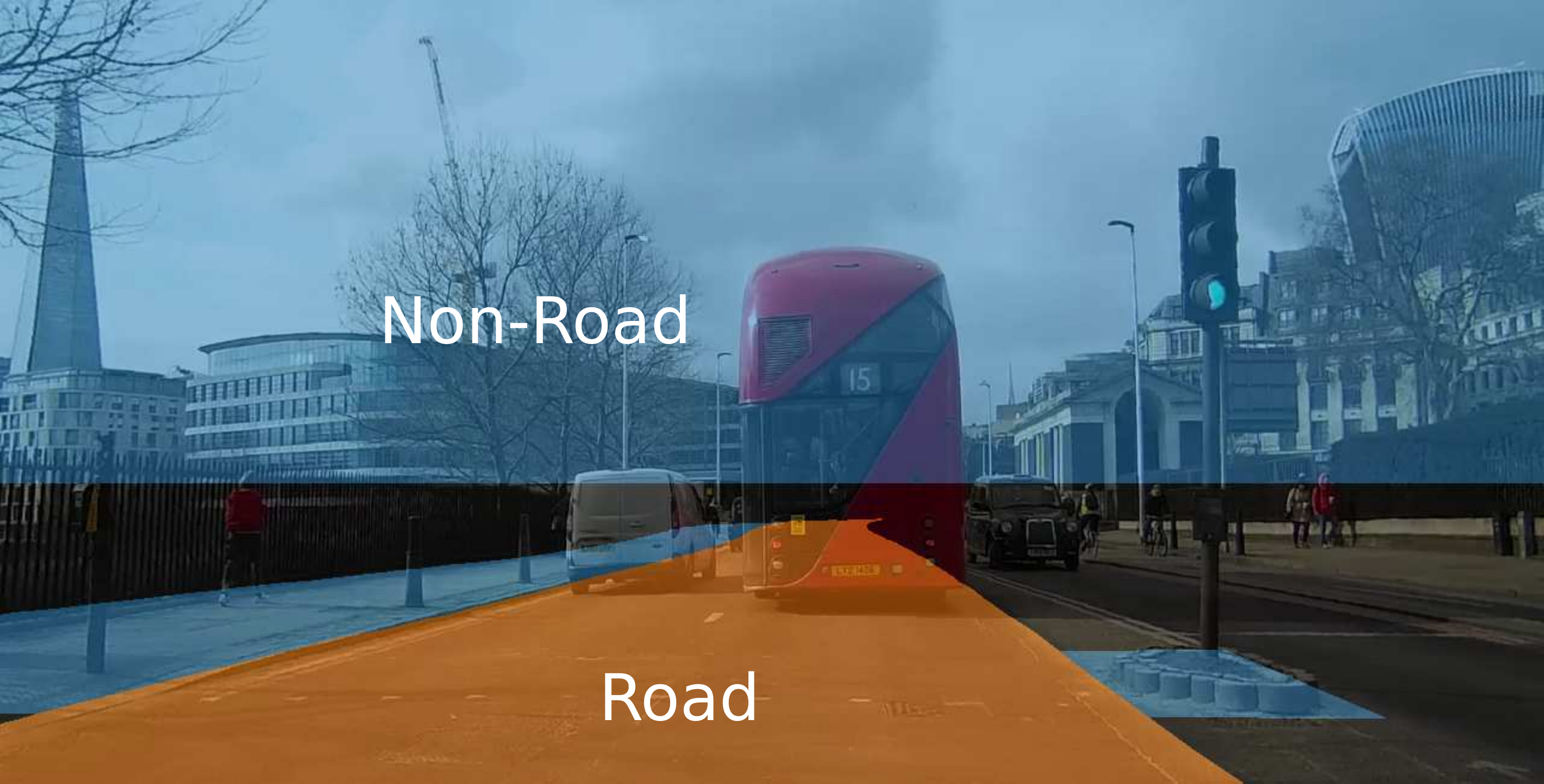}\\
	\includegraphics[width=0.45\linewidth]{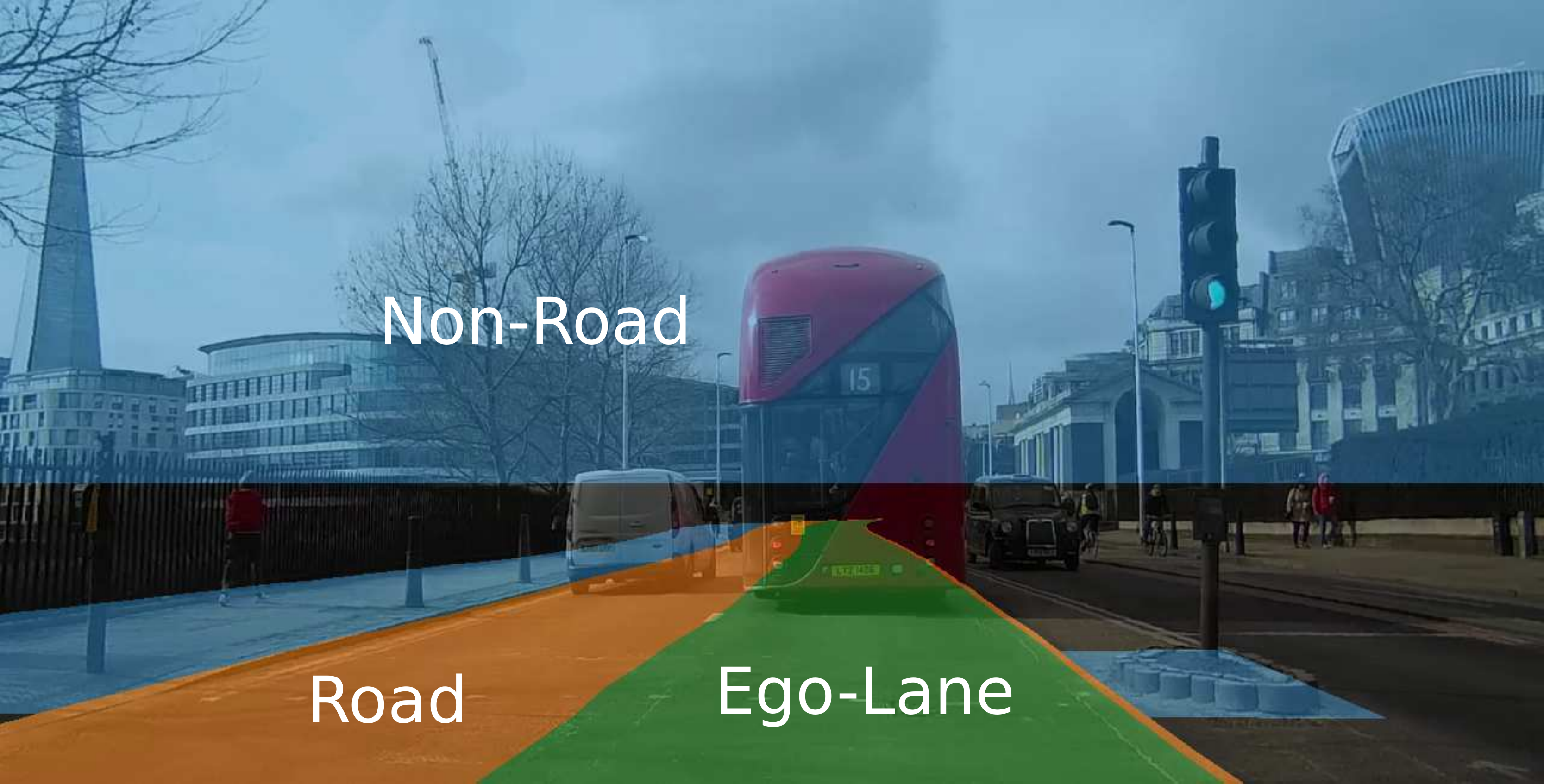}
	\includegraphics[width=0.45\linewidth]{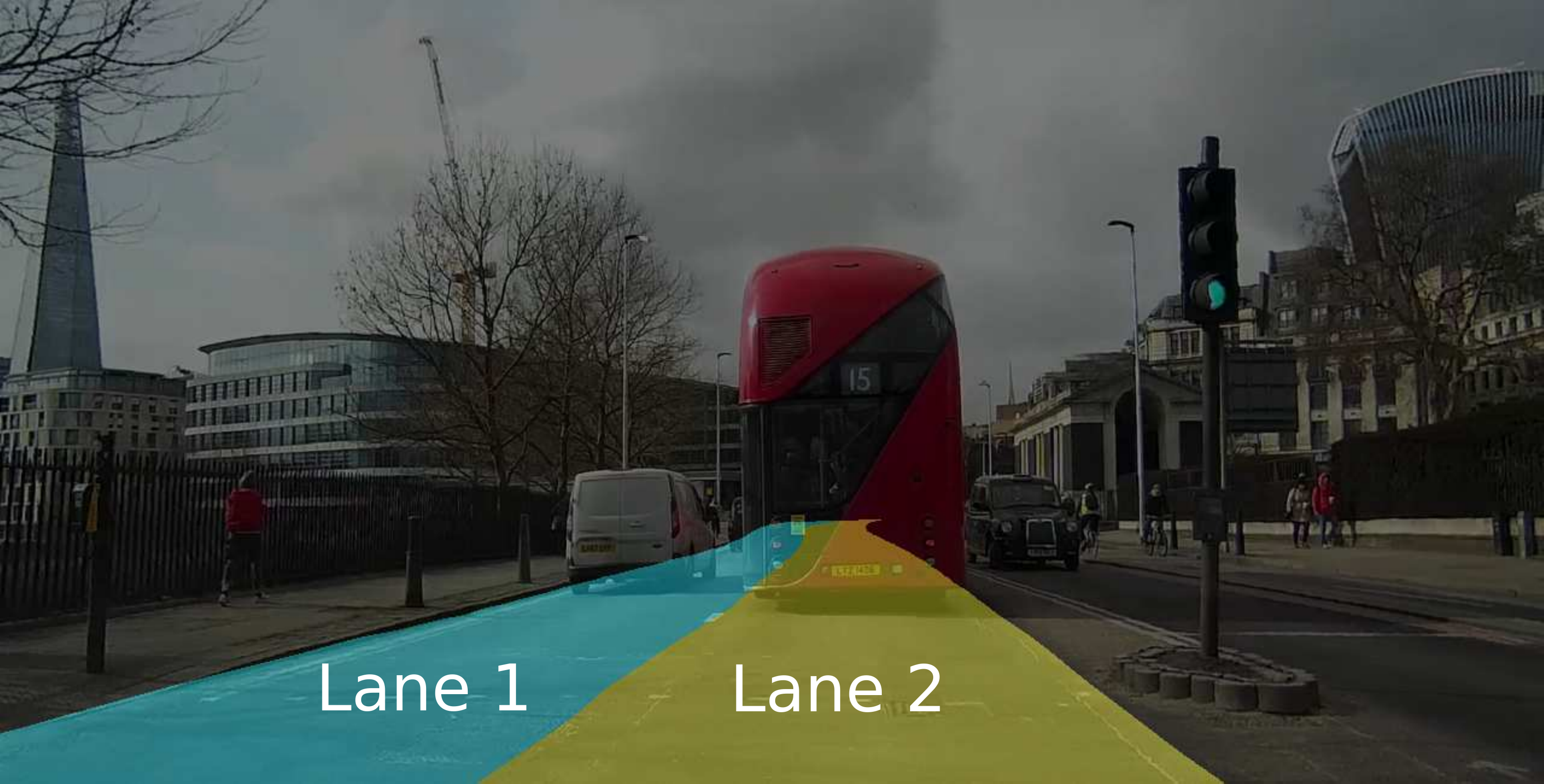}
	\caption{Example image from our dataset (top left), including annotations for road (top right), ego-lane (bottom left) and lane instance (bottom right). Road and lanes below vehicles are annotated despite being occluded. Non-coloured parts have not been annotated, i.e. the class is not known.}
	\label{fig:streetscene}
\end{figure}
% datasets
Various datasets include road area as a detection task, in addition to many other semantic segmentation classes \cite{brostow2008segmentation,brostow2009semantic,sengupta2012automatic,scharwachter2013efficient,cordts2016cityscapes,neuhold2017mapillary,matzen2013nyc3dcars}.
Some datasets also includes the ego-lane \cite{fritsch2013new}, which is useful for lane following tasks.
Few datasets provide lane instances \cite{aly2008real,TuSimple2017}, which are needed for more sophisticated driving manoeuvres.
Aly et. al. \cite{aly2008real} provide a relatively limited annotation of 4 single coordinates per lane border. TuSimple \cite{TuSimple2017} offer a large number of sequences, but for highway driving only.
Tab.~\ref{tab:dataset_comparison} provides an overview of the publicly available datasets. Our average annotation time per image is much lower. 
However, our provided classes are different, since we focus on lane instances (and thus ignore other semantic segmentation classes like vehicle, building, person, etc.). Furthermore, our data provides road surface annotations in dense traffic scenarios despite occlusions, i.e. we provide the road label below the vehicles (see Fig.~\ref{fig:streetscene}). This is different from typical semantic segmentation labels, which provide a label for the occluding object instead \cite{brostow2008segmentation,brostow2009semantic,sengupta2012automatic,scharwachter2013efficient,cordts2016cityscapes,neuhold2017mapillary}. 
Another approach to efficiently obtain labels is to create a virtual world where everything is known a-priori \cite{richter2016playing,ros2016synthia,Gaidon2016Virtual}. However, current methods do not reach the fidelity of real images.

%\begin{itemize}
%    \item Introduction
%    \item the problem
%    \item our idea
%\end{itemize}

%\section{Related Work}

\ctable[
caption = {Comparison of the available datasets.
Label time per image is only shown if provided by the authors.
Many datasets are not only targeting the road layout, and thus the labelling includes more classes.},
label = tab:dataset_comparison,
pos = htbp,
doinside = \scriptsize
]{lcrrccccr}{
    \tnote[a]{\scriptsize Only single images are annotated, but additional (non-annotated) image sequences are provided.}
    \tnote[b]{\scriptsize Road area is implicitly annotated by the given lanes.}
    \tnote[c]{\scriptsize Annotated ground instead of road, i.e. it includes non-drivable area.}
    \tnote[d]{\scriptsize Limited to three instances: ego-lane and left/right of ego-lane.}
}{
        & & \#labeled & & img. & road & ego & lane & label time \\

Name & Year & frames & \#videos & seq. & area & lane & instances & per img. \\

\toprule
Caltech Lanes \cite{aly2008real} & 2008 & 1,224 & 4 & \checkmark & \checkmark\tmark[b] & - & \checkmark & - \\

CamVid \cite{brostow2008segmentation,brostow2009semantic} & 2008 & 701 & 4 & \checkmark & \checkmark & - & - & 20 min \\

Yotta \cite{sengupta2012automatic} & 2012 & 86 & 1 & \checkmark & \checkmark & - & - & - \\

Daimler USD \cite{scharwachter2013efficient} & 2013 & 500 & - & - & \checkmark\tmark[c] & - & - & - \\

KITTI-Road \cite{fritsch2013new} & 2013 & 600 & - & - & \checkmark & \checkmark & - & - \\

NYC3DCars \cite{matzen2013nyc3dcars} & 2013 & 1,287 & - & - & \checkmark  & - & - & - \\

Cityscapes \cite{cordts2016cityscapes} (fine) & 2016 & 5,000 & - & \checkmark\tmark[a] & \checkmark & - & - & 90 min \\

Cityscapes \cite{cordts2016cityscapes} (coarse) & 2016 & 20,000 & - & \checkmark\tmark[a] & \checkmark & - & - & 7 min \\

Mapillary Vistas \cite{neuhold2017mapillary} & 2017 & 20,000 & - & - & \checkmark & - & - & 94 min \\

TuSimple \cite{TuSimple2017} & 2017 & 3,626 & 3,626 & \checkmark\tmark[a] & \checkmark\tmark[b] & \checkmark & \checkmark\tmark[d] & - \\

\textbf{Our Lanes} & 2018 & 23,980 & 402 & \checkmark & \checkmark & \checkmark & \checkmark & 5 sec \\

\bottomrule
}

%automated annotation methods:
Some previous work has aimed at creating semi-automated object detections in autonomous driving scenarios. \cite{leibe2007dynamic,matzen2013nyc3dcars} use structure-from-motion (SFM) to estimate the scene geometry and dynamic objects. \cite{borkar2012novel} proposes to annotate lanes in the birds-eye view and then back-project and interpolate the lane boundaries into the sequence of original camera images. \cite{laddha2016map} uses alignment with OpenStreetMap to generate ground-truth for the road. \cite{xie2016semantic} allows for bounding box annotations of Lidar point-clouds in 3D for road and other static scene components. These annotations are then back-projected to each camera image as semantic labels and they report a similar annotation speed-up as ours: 13.5 sec per image. \cite{barnes2016find} propose to detect and project the future driven path in images, without the focus of lane annotations. This means the path is not adapted to lane widths and crosses over lanes and junctions. Both \cite{xie2016semantic,barnes2016find} require an expensive sensor suite, which includes calibrated cameras and Lidar. In contrast, our method is applicable to data from a GPS enabled dash-cam. The overall contributions of this work include:
(1) The release of a new dataset for lane instance and road segmentation,
(2) A semi-automated annotation method for lane instances in 3D, requiring only inexpensive dash-cam equipment,
(3) Road surface annotations in dense traffic scenarios despite occlusion, and
(4) Experimental results for road, ego-lane and lane instance segmentation using a CNN.

\section{Video Collection}
Videos and associated GPS data were captured with a standard Nextbase 402G dashcam recording at a resolution of 1920x1080 at 30 frames per second and compressed with the H.264 standard.
% Note: Leave locations out for review
%Data was collected around Cambridge, London, Dartmoor and Yorkshire.
The camera was mounted on the inside of the car windscreen, roughly along the centre line of the vehicle and approximately aligned with the axis of motion. Fig.~\ref{fig:streetscene} (top left) shows an example image from our collected data. In order to remove parts where the car moves very slowly or stands still (which is common in urban environments), we only include frames that are at least 1m apart according to the GPS. Finally, we split the recorded data into sequences of 200m in length, since smaller sequences are easier to handle (e.g. no need for key-frame bundle adjustment, and faster loading times).

\section{Video Annotation}\label{sec:annotation}
%\begin{itemize}
%    \item which classes: road, ego-lane and lane instances
%    \item provide lane instances as mask? IDs should be consistent over time
%\end{itemize}
The initial annotation step is automated and provides an estimate of the road surface in 3D space, along with an estimate for the ego-lane (see Sec.~\ref{sec:annotation-automated}). Then the estimates are corrected manually and further annotations are added in the road surface space. The labels are then projected into the 2D camera views, allowing the annotation of all images in the sequence at once (see Sec.~\ref{sec:annotation-manual}).

\subsection{Automated Ego-lane Estimation in 3D}\label{sec:annotation-automated}
%\begin{itemize}
%    \item How does OpenSFM \cite{OpenSFM2014} work
%    \item calibrate camera: forward point, road normal and height within car
%    \item Project path back to images in 2D
%    \item Include sketch of how our method works
%\end{itemize}
Given a dash-cam video sequence of $N$ frames from a camera with unknown intrinsic and extrinsic parameters, the goal is to determine the road surface in 3D and project an estimate of the ego-lane onto this surface. 
To this end, we first apply OpenSfM \cite{OpenSFM2014}, a structure from motion algorithm, to obtain the 3D camera locations $\mathbf{c}_{i}$ and poses $\mathbf{R}_{i}$ for each frame $i\in\{1,...,N\}$ in a global coordinate system, as well as the camera projective transform $P(\cdot)$, which includes the estimated focal length and distortion parameters
($\mathbf{R}_{i} \in \mathbb{R}^{3 \times 3}$ are 3D rotation matrices).
OpenSfM reconstructions are not perfect, and failure cases are filtered during the manual annotation process.

\begin{figure}[htb]
    \centering
    \includegraphics[width=0.45\linewidth]{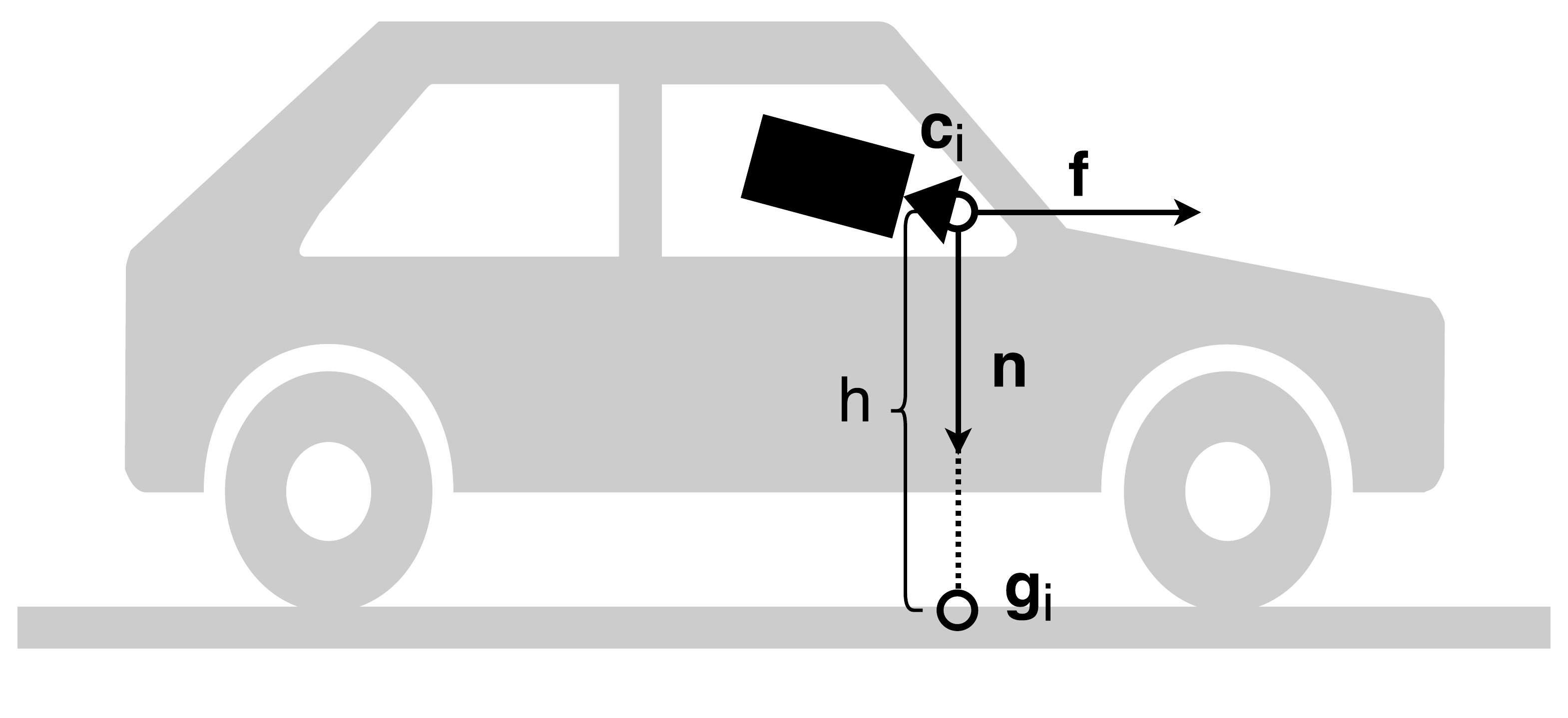}
    \includegraphics[width=0.45\linewidth]{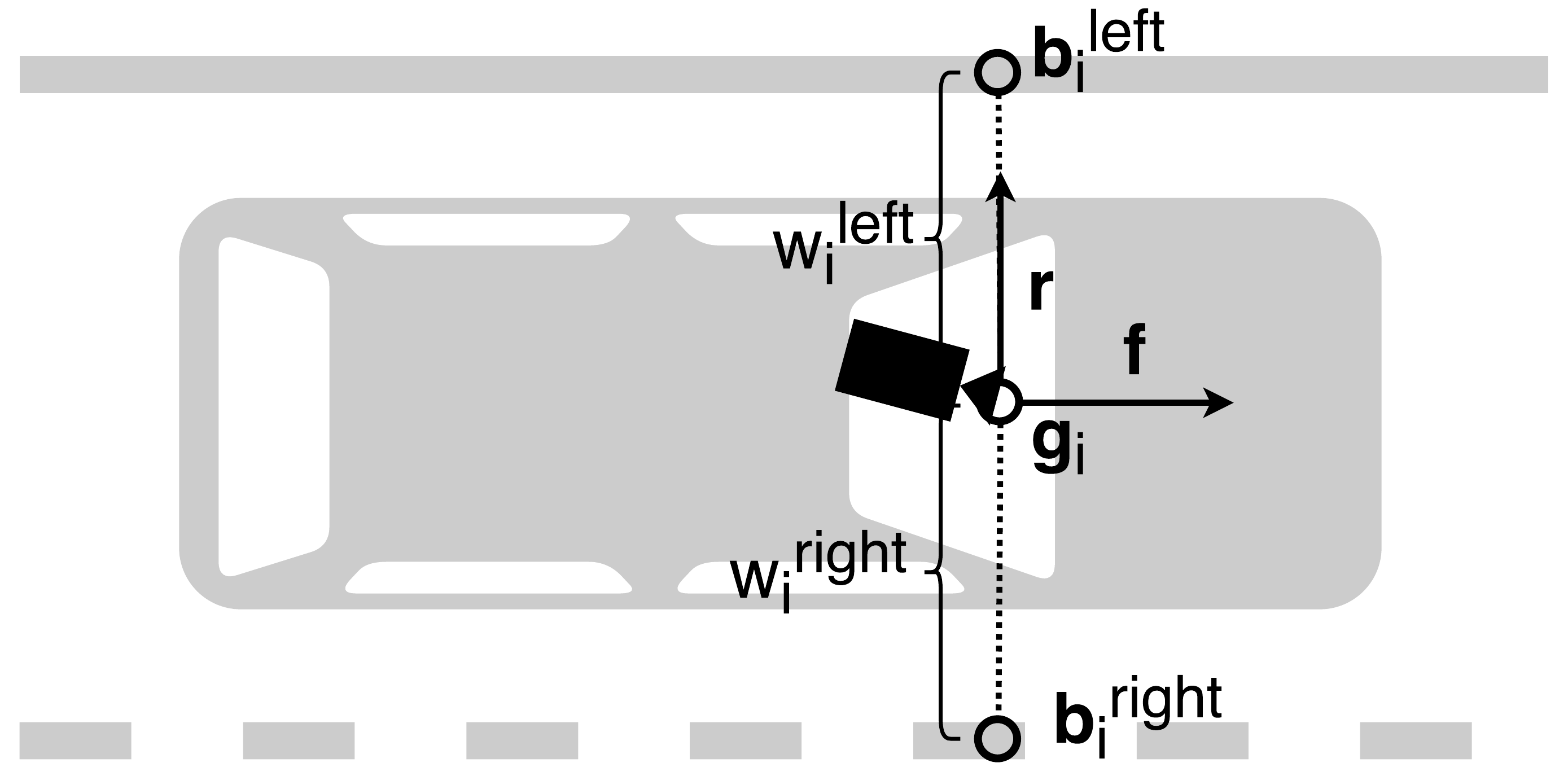}
    \caption{
        Estimation of the lane border points $\mathbf{b}_i^{left}, \mathbf{b}_i^{right}$ at frame $i$. 
        $\mathbf{c}_i$ is the camera position at frame $i$ (obtained via SfM),
        $\mathbf{g}_i$ is point on the road below the camera,
        $h$ is the height of the camera above the road,
        $\mathbf{f}$ is the forward direction,
        $\mathbf{n}$ is the normal vector of the road plane,
        $\mathbf{r}$ is the horizontal vector across the lane
        ($\mathbf{f}$, $\mathbf{n}$ and $\mathbf{r}$ are relative to the camera orientation) and
        $w_i^{left}, w_i^{right}$ are the distances to the left and right ego-lane borders.}
    \label{fig:borders_per_frame}
\end{figure}

We assume that the road is a 2D manifold embedded in the 3D world. The local curvature of the road is low, and thus the orientation of the vehicle wheels provide a good estimate of the local surface gradient. The camera is fixed within the vehicle with a static translation and rotation from the current road plane (i.e. we assume the vehicle body follows the road plane and neglect suspension movement). Thus the ground point $\mathbf{g}_{i}$ on the road below the camera at frame $i$ is calculated as $\mathbf{g}_{i}=\mathbf{c}_{i}+h\mathbf{R}_{i}\mathbf{n}$, where $h$ is the height of the camera above the road and $\mathbf{n}$ is the surface normal of the road relative to the camera (see Fig.~\ref{fig:borders_per_frame}, left). 
The left and right ego-lane borders $\mathbf{b}_{i}^{left}, \mathbf{b}_{i}^{right}$ can then be derived as
\begin{equation} \label{eq:border_points}
\begin{aligned}
\mathbf{b}_{i}^{left} & = \mathbf{g}_{i}+w_{i}^{left}\mathbf{R}_{i}\mathbf{r}\\
\mathbf{b}_{i}^{right} & = \mathbf{g}_{i}+w_{i}^{right}\mathbf{R}_{i}\mathbf{r}
\end{aligned}
\end{equation}
where $\mathbf{r}$ is the vector within the road plane, that is perpendicular to the driving direction and $w_{i}^{left}, w_{i}^{right}$ are the offsets to the left and right ego-lane borders. See Fig.~\ref{fig:borders_per_frame} (right) for an illustration. 
We make the simplifying assumption that the road surface is flat perpendicular to the direction of the car motion (but we don't assume that the road is flat generally - if our ego path travels over hills, this is captured in our ego path).

Given a frame $i$, we can project all future lane borders $\mathbf{b}_{j}$
($\mathbf{b}_{j}\in\{\mathbf{b}_{j}^{left},\mathbf{b}_{j}^{right}\}$ and $j>i$) into the local pixel coordinate system via
\begin{equation} \label{eq:future_border}
\hat{\mathbf{b}_{j}}=P\left(\mathbf{R}_{i}^{-1}(\mathbf{b}_{j}-\mathbf{c}_{i})\right)
\end{equation}
where $P()$ is the camera perspective transform obtained via OpenSfM \cite{OpenSFM2014}, that projects a 3D point in camera coordinates to a 2D pixel location in the image.
Then the lane annotations can be drawn as polygons of neighbouring future frames, i.e. with the corner points $\hat{\mathbf{b}}_{j}^{left},\hat{\mathbf{b}}_{j}^{right},\hat{\mathbf{b}}_{j+1}^{right},\hat{\mathbf{b}}_{j+1}^{left}$. 
This makes implicitly the assumption that the lane is piece-wise straight and flat between captured images. 
In the following part, we describe how to get the quantities $h$, $\mathbf{n}$, $\mathbf{r}$, $w_{i}^{left}$ and $w_{i}^{right}$. 
Note that $h$, $\mathbf{n}$ and $\mathbf{r}$ only need to be estimated once for all sequences with the same camera position. 

% camera height above road
The camera height above the road $h$ is easy to measure manually. However, in case this cannot be done (e.g. for dash-cam videos downloaded from the web) it is also possible to obtain the height of the camera using the estimated mesh of the road surface obtained from OpenSfM. A rough estimate for $h$ is sufficient, since it is corrected via manual annotation, see the following section.

\begin{figure}[htb]
    \centering
    \includegraphics[width=0.4\linewidth]{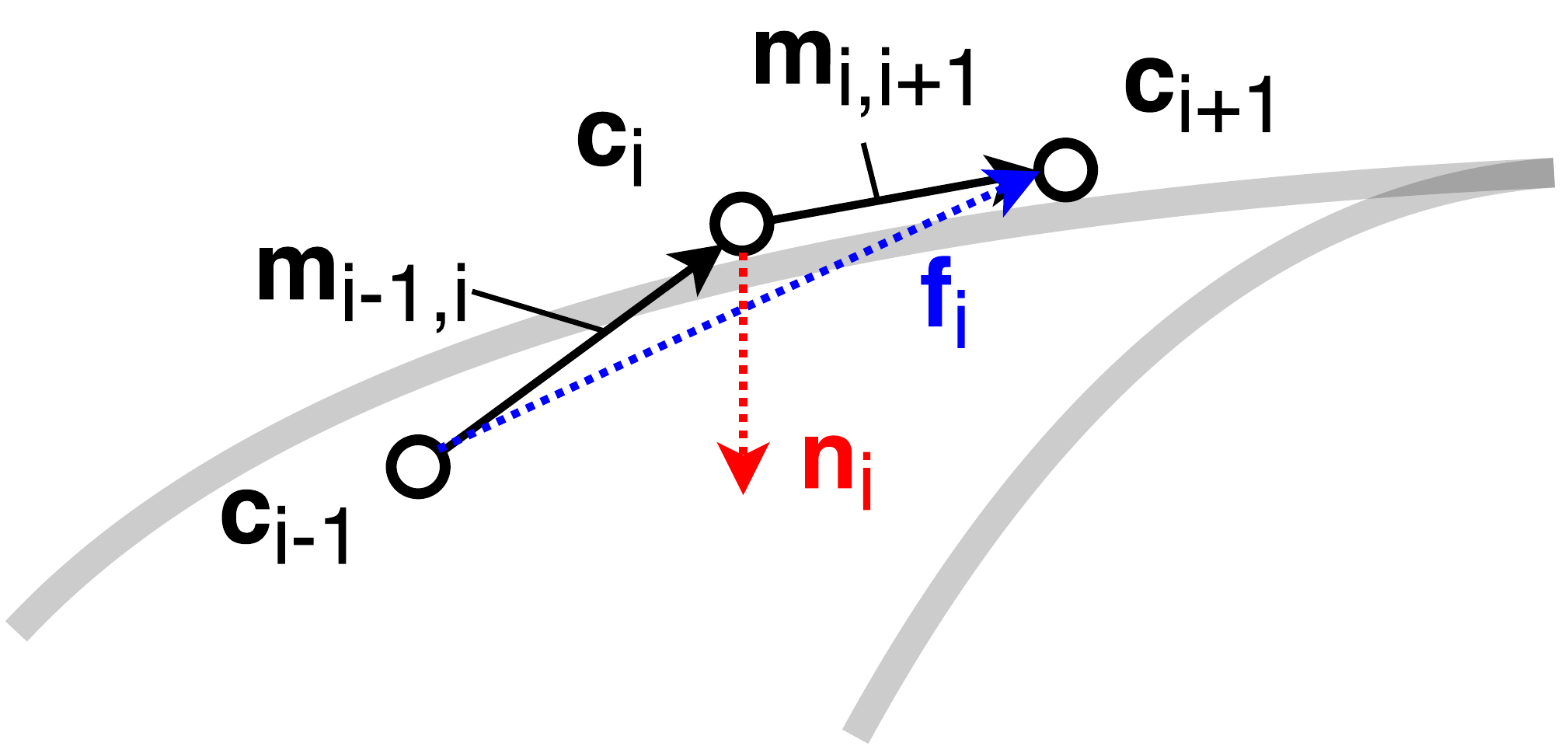}
    \caption{Estimation of the road normal $\mathbf{n}_i$ and forward direction $\mathbf{f}_i$ at a single frame $i$. The final estimate is an aggregate over all frames.}
    \label{fig:forward_and_normal}
\end{figure}

% road normal
In order to estimate the road normal $\mathbf{n}$, we use the fact that when the car moves around a turn, the vectors representing it's motion $\mathbf{m}$ will all lie in the road plane, and thus taking the cross product of them will result in the road normal, see Fig.~\ref{fig:forward_and_normal}. Let $\mathbf{m}_{i,j}$ be the normalised motion vector between frames $i$ and $j$, i.e.  $\mathbf{m}_{i,j} = \frac{\mathbf{c}_{j}-\mathbf{c}_{i}}{\left\Vert \mathbf{c}_{j}-\mathbf{c}_{i}\right\Vert }$. The estimated road normal at frame $i$ (in camera coordinates) is $\mathbf{n}_{i}=\mathbf{R}_{i}^{-1}(\mathbf{m}_{i-1,i}\otimes\mathbf{m}_{i,i+1})$, where $\otimes$ denotes the cross-product (see Fig.~\ref{fig:forward_and_normal}). The quality of this estimate depends highly on the degree of our previous assumptions being correct. To get a more reliable estimate, we average all $\mathbf{n}_{i}$ across the journey, and weight them  implicitly by the magnitude of the cross product:
\begin{equation} \label{eq:normal}
\mathbf{n}=\frac{1}{\sum_{i=2}^{N-2}\left\Vert \mathbf{n}_{i}\right\Vert }\sum_{i=2}^{N-2}\mathbf{n}_{i}
\end{equation}
We can only estimate the normal during turns, and thus this weighting scheme emphasises tight turns and ignores straight parts of the journey.
% vector in the road
$\mathbf{r}$ is perpendicular to the forward direction $\mathbf{f}$ and within the road plane, thus
\begin{equation} \label{eq:across_road}
\mathbf{r}=\mathbf{f}\otimes\mathbf{n}
\end{equation}
% forward point
The only quantity left is $\mathbf{f}$, which can be derived by using the fact that $\mathbf{m}_{i-1,i+1}$ is approximately parallel to the tangent at $\mathbf{c}_{i}$, if the rate of turn is low. Thus we can estimate the forward point at frame $i$ via $\mathbf{f}_{i}=\mathbf{R}_{i}^{-1}\mathbf{m}_{i-1,i+1}$, see Fig.~\ref{fig:forward_and_normal}.
As for the normal, we average all $\mathbf{f}_{i}$ over the journey to get a more reliable estimate:
\begin{align} \label{eq:forward}
\mathbf{f}&=\frac{1}{\sum_{i}a_{i}}\sum_{i=2}^{N-2}a_{i}\mathbf{f}_{i}\\
a_{i}&=\max(\mathbf{m}_{i-1,i}^{\top}\mathbf{m}_{i,i+1},0)
\end{align}
In this case, we weight the movements according the inner product $a_{i}$ in order to up-weight parts with a low rate of turn, while the $\max$ assures forward movement.

% road width w
$w_{i}^{left}$ and $w_{i}^{right}$ are crucial quantities to get the correct alignment of the annotated lane borders with the visible boundary, however automatic detection is non-trivial. Therefore we assume initially that the ego-lane has a fixed width $w$ and the car has travelled exactly in the centre, i.e. $w_{i}^{left}=\frac{1}{2}w$ and $w_{i}^{right}=-\frac{1}{2}w$ are both constant for all frames. Later (see the following section), we relax this assumption and get an improved estimate through manual annotation.
 
\begin{figure}[htb]
    \centering
    \includegraphics[width=0.8\linewidth]{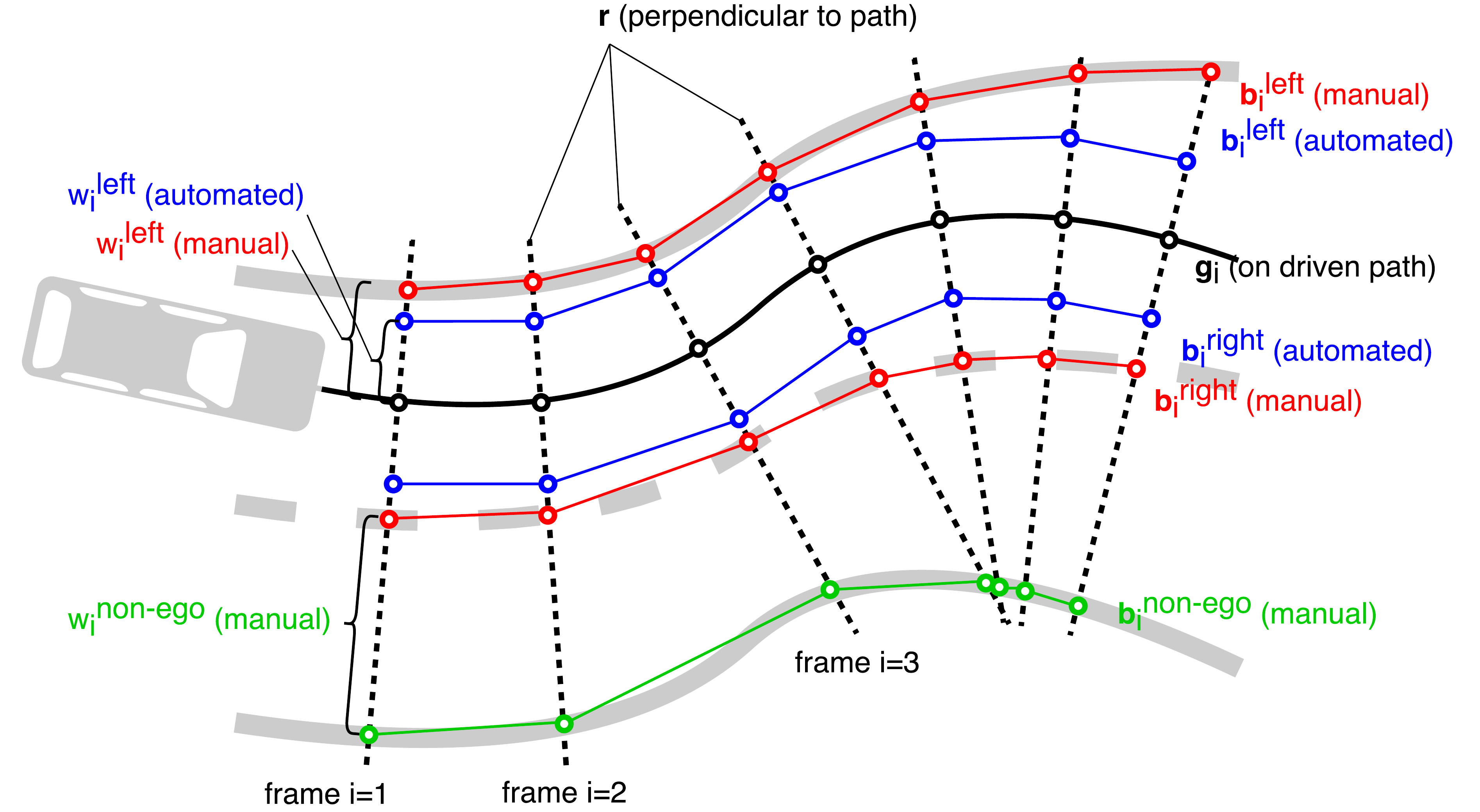}
    \caption{Visualisation of the lane estimates, seen from above. The automated estimate is shown in blue, while the manual correction is shown in red. A manually annotated additional lane is shown in green.  Initially, all $w_{i}^{left}$ and  $w_{i}^{right}$ are set to a constant value, and thus the estimate is parallel the driven path, which only approximately follows the true lane borders (in blue). Then the annotators can correct $w_{i}^{left}$ and  $w_{i}^{right}$ for each frame, which moves the border points along $\mathbf{r}$ (shown as dotted black line) until they align with the true border (shown in red). 
    Furthermore, annotators can add additional (non-ego) lanes and adjust their width $w_{i}^{non-ego}$.}
    \label{fig:egolane_estimation}
\end{figure}

In practice, we select a sequence with a lot of turns within the road plane to estimate $\mathbf{n}$ and a straight sequence to estimate $\mathbf{f}$. 
Then the same values are re-used for all sequences with the same static camera position. 
We only annotate the first part of the sequence, up until 100m from the end. We do this to avoid partial annotations on the final frames of a sequence which result from too few lane border points remaining ahead of a given frame.
A summary of the automated ego-lane annotation procedure is provided in Algorithm~\ref{alg:egolane} and a visualisation of the automated border point estimation is shown in Fig.~\ref{fig:egolane_estimation} (in blue).
\begin{algorithm}
    \caption{Automated ego-lane estimation}\label{alg:egolane}
    \begin{algorithmic}[1]
        \State Measure height of the camera above road $h$
        \State Apply OpenSFM to get $\mathbf{c}_{i},\mathbf{R}_{i}$
        \State Estimate road normal $\mathbf{n}$ according Eq.~(\ref{eq:normal})
        \State Estimate forward direction $\mathbf{f}$ according Eq.~(\ref{eq:forward})
        \State Derive vector across road $\mathbf{r}$ according Eq.~(\ref{eq:across_road})
        \State Set $w_{i}^{left}=\frac{1}{2}w$ and $w_{i}^{right}=-\frac{1}{2}w$, where $w$ is the default lane width
        \State Derive border points $\mathbf{b}_{i}^{left},\mathbf{b}_{i}^{right}$
        according Eq.~(\ref{eq:border_points})
        \For{each frame $i$}
            \State Get all future border points $\hat{\mathbf{b}}_{j}^{left},\hat{\mathbf{b}}_{j}^{right}$, $j>i$ according Eq.~(\ref{eq:future_border})
            \State Draw polygons with edges $\hat{\mathbf{b}}_{j}^{left},\hat{\mathbf{b}}_{j}^{right},\hat{\mathbf{b}}_{j+1}^{right},\hat{\mathbf{b}}_{j+1}^{left}$
        \EndFor
    \end{algorithmic}
\end{algorithm}

\subsection{Manual corrections and additional annotations}\label{sec:annotation-manual}
%\begin{itemize}
%    \item screenshot of annotator interface (with explanations)
%    \item measured time needed for the annotations (important!)
%\end{itemize}
Manual annotations serve three goals:
(1) exclude erroneous OpenSfM reconstructions
(2) to improve the automated estimate for the ego-lane,
(3) annotate additional lanes left and right of the ego-lane and
(4) annotate non-road areas.

\begin{figure}[tb]
\hspace*{\fill}%
	\begin{minipage}[t]{0.49\linewidth}
		\vspace{0pt}
		\includegraphics[width=\linewidth]{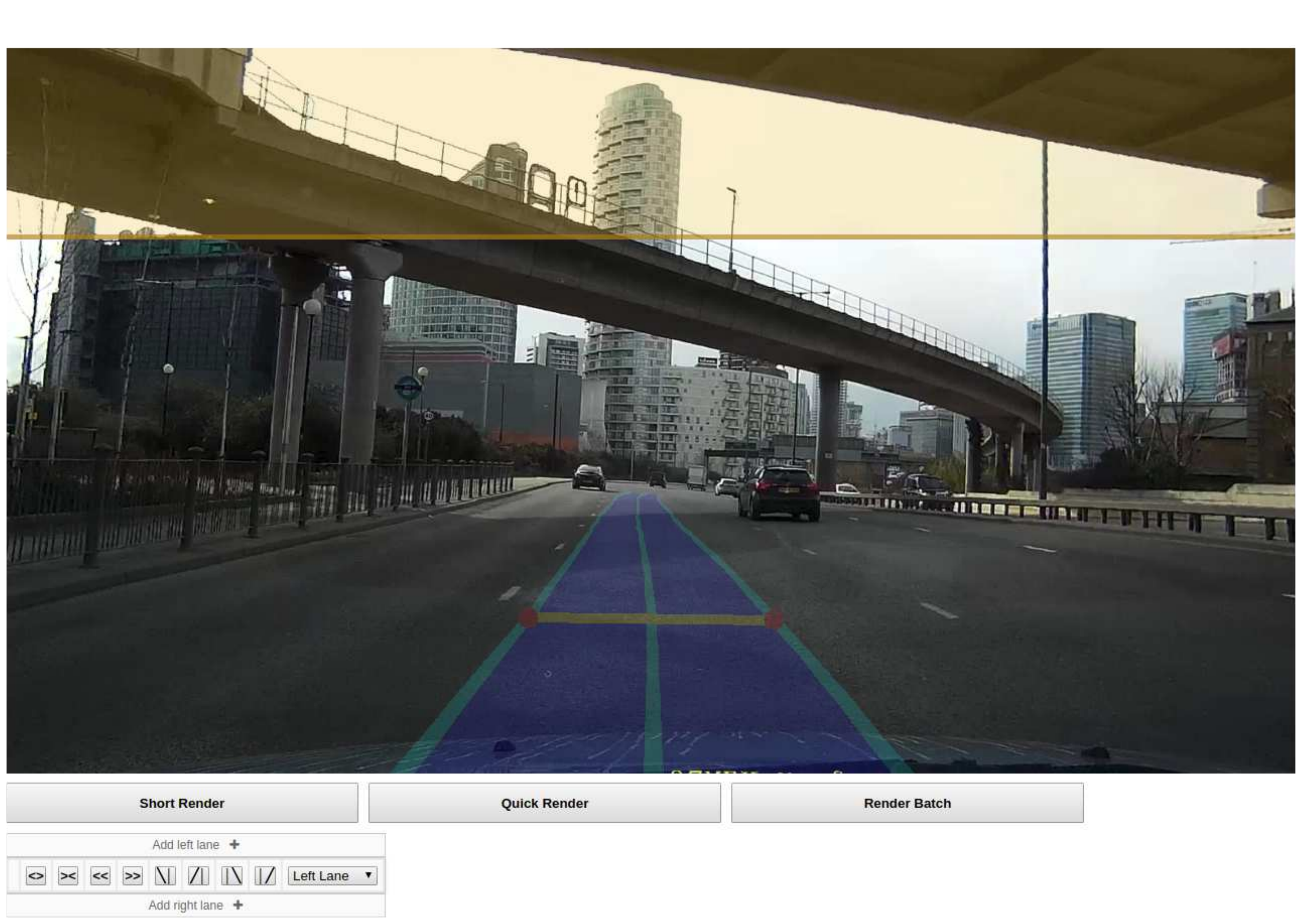}
	\end{minipage}
\hfill
	\begin{minipage}[t]{0.49\linewidth}
		\vspace{0pt}
		\includegraphics[width=\linewidth]{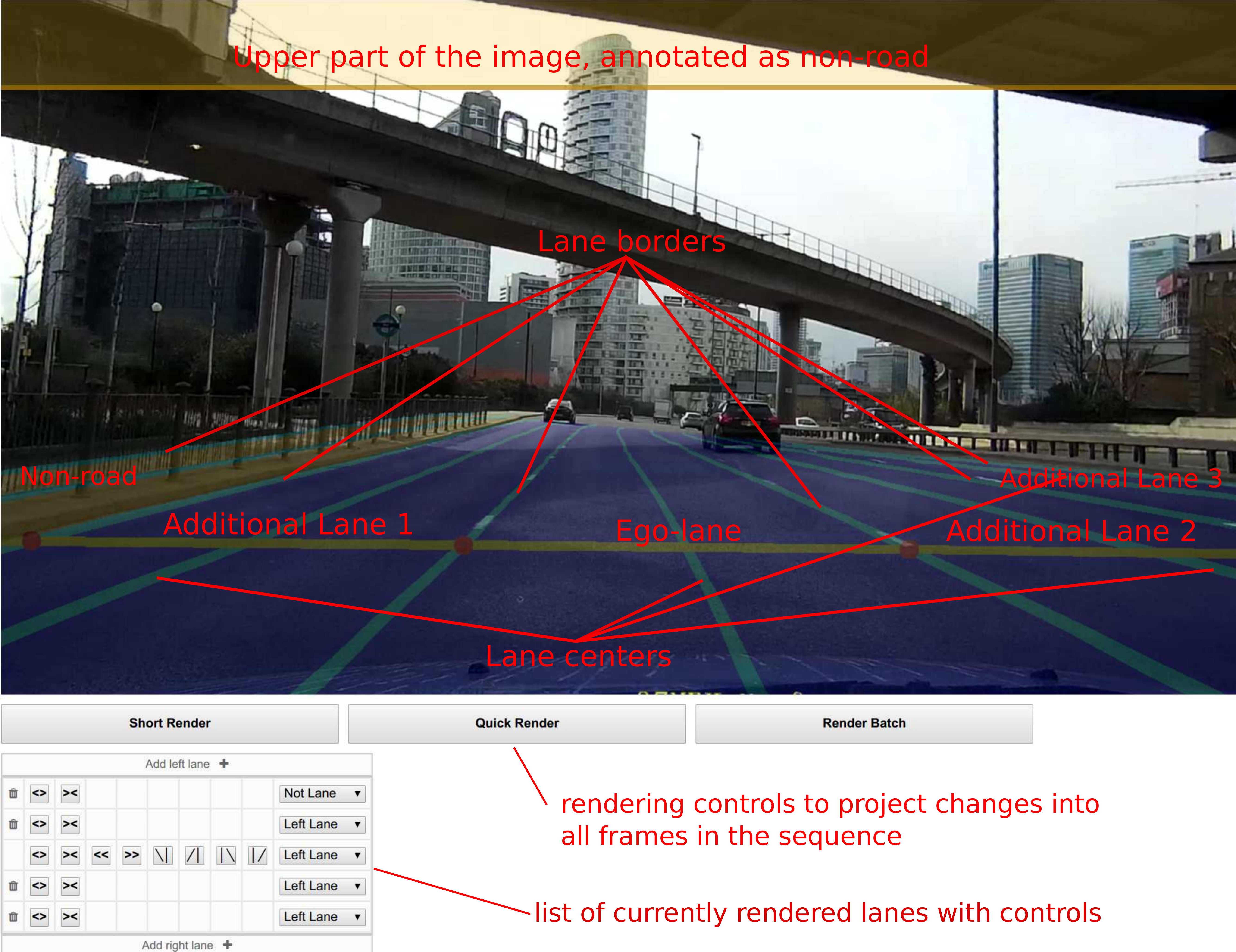}
	\end{minipage}
\hspace*{\fill}

    \caption{Annotator interface with the automated initial ego-lane estimate, given by the future driven path (left) and after manual corrections and additional lane annotations (right). The red text and lines are overlaid descriptions, all other colours are part of the interface.}
    \label{fig:interface}
\end{figure}

OpenSfM failures happened a few times, but they are easy to spot by the
annotator and subsequently excluded from the dataset.
In order to improve the ego-lane positions, the annotators are provided with a convenient interface to edit $h$, $w_{i}^{left}$ and $w_{i}^{right}$. Note that these quantities are only scalars (in contrast to 3D points), and are thus easily adjusted via keyboard input. We provide a live rendered view at a particular frame (see Fig.~\ref{fig:interface}, left), and immediate feedback is provided after changes. Also, it is easy to move forward or backward in the sequence. For improving the ego-lane, the annotators have the options to:
\begin{enumerate}
    \item Adjust $h$ (applies to the whole sequence)
    \item Adjust all $w_{i}^{left}$ or all $w_{i}^{right}$ (applies to the whole sequence)
    \item Adjust all $w_{j}^{left}$ or all $w_{j}^{right}$ from the current frame $i$ on, $j>i$ (applies to all future frames, relative to the current view)
\end{enumerate}
In order to keep the interface complexity low, only one scalar is edited at a time.
We observed that during a typical drive, the car is moving parallel to the ego-lane \emph{most of the time}. Also, lanes have a constant width \emph{most of the time}.
If both holds, then it is sufficient to use (2) to edit the lane borders for the whole sequence. Only in the case that the car deviates from the parallel path, or the lane width changes, the annotator needs option (3).

New lanes can be placed adjacent to current ones by a simple button click. This generates a new sequence of $\mathbf{b}_{i}^{non-ego}$, either on the left or right of the current lanes (see \ref{fig:egolane_estimation}). As for the ego-lane, the annotator can adjust the corresponding $w_{i}^{non-ego}$.  Equivalently, a non-road surface can be added next to current lanes, in the same way as if it were a lane, i.e. by getting its own set of $\mathbf{b}_{i}^{non-ego}$ and $w_{i}^{non-ego}$. In addition to that, a fixed part on top of the image can be annotated with non-road, as the road is usually found in the lower part of the image (except for very hilly regions or extreme camera angles).

Fig.~\ref{fig:interface} (left) shows the interface used by the annotators. In the centre of the image, the ego-path can be seen projected into this frame. In the bottom-left, the annotator is provided with controls to manipulate rendered lanes (narrow, widen, move to the left or right, move the boundaries of the lane etc.) and add new lanes. In the top right of the screen (not visible), the annotator is provided with the means to adjust the camera height, to match the reconstruction to the road surface, and the crop height, to exclude the vehicle’s dash or bonnet. All annotations are performed in the estimated 3D road plane, but immediate feedback is provided via projection in the 2D camera view. The annotator can easily skip forward and backward in the sequence to determine if the labels align with the image, and correct them if needed. An example of a corrected sequence is shown in Fig.~\ref{fig:egolane_estimation} (in red). Fig.~\ref{fig:streetscene} shows an example of the rendered annotations and the supplementary material contains an example video. 

\label{sec:statistics}
\section{Dataset Statistics and Split}
% commands used on ml6: 
%   cd /data_ssd/static_scene_perception_datasets/brooklane_dataset_paper_friday
%   # For all sequence count
%   find `pwd`/road/h5/{test,train,val} -type d | grep -E 'batch_[0-9]+$'| wc -l 
%   # Number of training sequences
%   find `pwd`/road/h5/{train,val} -type d | grep -E 'batch_[0-9]+$'| wc -l 
%   # Number of test sequences
%   find `pwd`/road/h5/test -type d | grep -E 'batch_[0-9]+$'| wc -l 
%   # For all image count
%   find `pwd`/road/h5/{test,train,val} -type f | grep originals_cropped | wc -l 
%   # Number of training images
%   find `pwd`/road/h5/{train,val} -type f | grep originals_cropped | wc -l 
%   # Number of test images
%   find `pwd`/road/h5/test -type f | grep originals_cropped | wc -l 
The full annotated set includes $402$ sequences, $23,979$ images in total, and thus on average 60 images per sequence. Tab.~\ref{tab:statistics_annotations} shows a breakdown of the included annotation types. In total, there were 47,497 lane instances annotated, i.e. 118.2 per sequence. Instance IDs are consistent across a sequence, i.e. consecutive frames will use the same instance ID for the same lane. Furthermore, the annotators have been instructed to categorise each sequence according the scene type: urban, highway or rural. The breakdown of the sequences is shown in Tab.~\ref{tab:statistics_scene}.
We plan to update the dataset with new sequences, once they become available.

\begin{table}[tb]
    \centering
    \caption{Dataset breakdown according scene type (a) and annotation coverage (b). Coverage of scene types and instances is measured as percentage of the total number of sequences, while the coverage of annotations is measured as percentage of the total number of pixels.}
    \label{tab:statistics}
\subfloat[]{%
	\begin{minipage}[t]{.3\linewidth}%
			\centering%
	    \begin{tabular}[t]{lr}
	        \toprule
	        Scene type & \\
	        \midrule
	        Urban & 58.61\% \\
	        Highway & 10.56\% \\
	        Rural & 30.83\% \\
	        \bottomrule
	    \end{tabular}%
	    \label{tab:statistics_scene}%
	\end{minipage}%
}%
\subfloat[]{%
	\begin{minipage}[t]{.6\linewidth}%
		\centering%
	    \begin{tabular}[t]{lr}
	        \toprule
	        Annotation type & \\
	        \midrule
	        annotation density & 77.53\% \\
	        non-road & 62.13\% \\
	        road & 15.40\% \\
	        ego-lane & 8.84\% \\
	        \midrule
	        & mean/median/min/max \\
	        \midrule
	        \#instances (per sequence) & 2.2/2/1/6 \\
	        \bottomrule
	    \end{tabular}%
	    \label{tab:statistics_annotations}%
	\end{minipage}%
}%
\end{table}

We split the data into two sets, for training and testing. 
The train set comprises 360 sequences and a total of $21,355$ frames, while the test set includes $42$ sequences and $2,624$ frames. The test set was selected to include the same urban/motorway/rural distribution as the train set. 
The frames of the training set are made available\footnote{online at \url{https://five.ai/datasets}} with both images and annotations while only the images are provided for the testing set.

Furthermore, we have measured the average annotation time per scene type, and find that there is a large variation, with an urban scene taking roughly 3 times longer than a highway or countryside scene of similar length (see Tab.~\ref{tab:annotation_time}).
This is due to the varying complexity in terms of the road layout, which is caused by various factors:
the frequency of junctions and side roads, overall complexity of lane structure and additional features such as traffic islands and cycle lanes that are typically not found outside of an urban setting.
\begin{table}[htb]
\begin{minipage}[t]{0.4\linewidth}
    \centering
    \caption{Average annotation time in seconds.}
    \label{tab:annotation_time}
    \begin{tabular}{lrrr}
        \toprule
        Scene type & Urban & Highway & Rural \\
        \midrule
        Per sequence & 361 & 100 & 140 \\
        Per image & 5 & 2 & 2 \\
        \bottomrule
    \end{tabular}
	\end{minipage}
	\begin{minipage}[t]{0.6\linewidth}
	\centering
    \caption{Agreement of the annotators}
	\label{tab:annotator_agreement}
	\begin{tabular}{lrr}
		\toprule
		Task & IoU & std\\
		\midrule
		Road vs non-road & 97.2 & $\pm$ 1.5 \\
		Ego vs road vs non-road & 94.3 & $\pm$ 3.4 \\
		\midrule
		& AP@50 & AP\\
		\midrule
		Lane instance segmentation & 99.0 & 84.4 \\
		\bottomrule
	\end{tabular}
\end{minipage}
\end{table}

The annotation quality is measured through agreement between the two annotators on 12 randomly selected sequences. 84.3\% of the pixels have been given a label by at least 1 annotator, with 67.3\% of these being given an annotation by both annotators; i.e. 56.8\% of all pixels were given an annotation by both annotators. We measure the agreement on these overlapping labels via Intersection-over-Union (IoU) and agreement of instances using Average Precision (AP) and AP@50 (average precision with instance IoU greater than 50\%). The results are shown in Tab.~\ref{tab:annotator_agreement}. The standard deviation is calculated over the 12 sequences.
\begin{table}[htb]
    
\end{table}

\section{Experiments}\label{sec:results}
%TODO: check contents
%\begin{itemize}
%    \item describe a simple CNN (possible: U-net, Mask-RCNN)
%    \item evaluation procedure, use IoU (same evaluation as for annotators)
%    \item optional: cross-dataset experiments (e.g. train on ours, test on kitti)
%    \item experiment 1: road/non-road
%    \item experiment 2: ego-lane/road-other/non-road
%    \item experiment 3: lane instance segmentation
%\end{itemize}

To demonstrate the results achievable using our annotations we present evaluation procedures, models and results for two example tasks: semantic segmentation of the road and ego-lane, as well as lane instance segmentation. 

\subsection{Road and Ego-Lane Segmentation}

The labels and data described in~\ref{sec:statistics} directly allow for two segmentation tasks: Road/Non-Road detection (ROAD) and Ego/Non-Ego/Non-Road lane detection (EGO).
For our baseline we used the well studied SegNet~\cite{badrinarayanan2015segnet} architecture, trained independently for both the EGO and ROAD experiments.
In addition to an evaluation on our data, we provide ROAD and EGO cross-database results for CityScapes (fine), Mapillary and KITTI Lanes. We have selected a simple baseline model and thus the overall results are lower than those reported for models tailored to the respective datasets, as can be seen in the leaderboards of CityScapes, Mapillary and KITTI. Thus our results should not be seen as an upper performance limit. Nevertheless, we deem them a good indicator on how models generalise across datasets.

For each dataset, we use $10\%$ of training sequences for validation.
During training, we pre-process each input image by resizing it to have a height of 330px and extracting a random crop of $320 \times 320$px. We use the ADAM  optimiser \cite{Kingma14Method} with a learning rate of 0.001 which we decay to 0.0005 after $25,000$ steps and then to 0.0001 after $50,000$ steps. We trained for $100,000$ training steps, and select the model with the best validation loss. Our mini batch size was 2 and the optimisation was performed on a per pixel cross entropy loss.

We train one separate model per dataset and per task. This leads to 4 models for ROAD, trained on our data, CityScapes (fine), Mapillary and KITTI Lanes. EGO labels are only available for the UM portion of KITTI Lanes and our data, hence we train 2 models for EGO. 

For each model we report the IoU, and additionally the F1 score as it is the default for KITTI. We measure each model on held out data from every dataset. For CityScapes and Mapillary the held out sets are their respective pre-defined validation sets, for our dataset the held out set is our test set (as defined in Sec.~\ref{sec:statistics}). The exception to this scheme is KITTI Lanes which is very small and has no available annotated held out set. Therefore we use the entire set for training the KITTI model, and the same set for the evaluation of other models. We report the average IoU and F1 across classes for each task. Note that we cropped the car hood and ornament from the CityScapes data, since it is not present in other datasets (otherwise the results drop significantly). It should also be noted that the results are not directly comparable to the intended evaluation of CityScapes, Mapillary or KITTI Lanes due to the different treatment of the road occluded by vehicles.

The ROAD results are shown in Tab.~\ref{tab:results_segmentation_road_non_road} and the EGO results in Tab.~\ref{tab:results_segmentation_ego_non_ego_non_road}. First, we note that IoU and F1 follow the same trend, while F1 is a bit larger in absolute values.
We see a clear trend between the datasets. Firstly, the highest IoUs are achieved when training and testing subsets are from the same data. This points to an overall generalisation issue; no dataset (including our own) achieves the same performance on other data. 
The model trained on KITTI shows the worst cross-dataset average. This is not surprising, since it is also the smallest set (it contains only 289 images for the ROAD task and 95 images for the EGO task). Cityscapes does better, but there is still a bigger gap to ours and Mapillary, probably due to lower diversity.
Mapillary is similar to ours in size and achieves almost the same performance. The slightly lower results could be due to its different viewpoints, since it contains images taken from non-road perspectives, e.g. side-walks.

\begin{table}
    \centering
    \caption{Results for the ROAD task, measured by IoU and F1 score. Off-diagonal results are from cross-dataset experiments. The column determines which set the model was trained on, and the row determines the source of the evaluation set. The reported column average includes only cross-dataset experiments.}
    \label{tab:results_segmentation_road_non_road}
\begin{tabular}{ll|p{1.6cm}p{1.6cm}p{1.6cm}p{1.6cm}}
	\multicolumn{2}{l}{\multirow{2}{*}{IoU}}       & \multicolumn{4}{l}{Trained On}               \\
	\multicolumn{2}{l}{}                           & Ours  & Mapillary & CityScapes & KITTI  \\
	\midrule
	\multirow{4}{*}{\rotatebox[origin=c]{90}{Tested On}} & Our Test Set      & 95.0 & 85.4     & 73.2     & 71.0       \\
	& Mapillary Val     & 82.9 & 90.0     & 79.6      & 69.6 \\
	& CityScapes Val    & 85.2 & 85.2    & 90.0      & 60.4      \\
	& KITTI Train & 83.8 & 72.6     & 74.6      & -          \\
	\midrule
	& Cross-dataset Average  & 84.0 & 81.1    & 75.8      & 67.0         \\
	\midrule
\end{tabular}
\begin{tabular}{ll|p{1.6cm}p{1.6cm}p{1.6cm}p{1.6cm}}
	\multicolumn{2}{l}{\multirow{2}{*}{F1}}       & \multicolumn{4}{l}{Trained On}               \\
	\multicolumn{2}{l}{}                           & Ours  & Mapillary & CityScapes & KITTI  \\
	\midrule
	\multirow{4}{*}{\rotatebox[origin=c]{90}{Tested On}} & Our Test Set      & 97.4 & 91.9     & 83.7      & 81.6       \\
	& Mapillary Val     & 90.4 & 94.7     & 88.3      & 81.0       \\
	& CityScapes Val    & 91.9 & 91.9    & 94.7      & 74.0      \\
	& KITTI Train & 90.9 & 83.5     & 84.8      & -          \\
	\midrule
	& Cross-dataset Average & 91.1 & 89.1     & 85.6      & 75.8          \\
	\bottomrule
\end{tabular}
\end{table}

\begin{table}
	\begin{minipage}[t]{.7\linewidth}
		\centering
		\caption{Results for the EGO task, measured by IoU and F1 score.}
		\label{tab:results_segmentation_ego_non_ego_non_road}
        \begin{tabular}{lrrr}
            \toprule
            Train & Test & \hspace{1cm}IoU & \hspace{0.5cm}F1 \\
            \midrule
            Ours & Ours & 88.5 & 93.7 \\
            Ours & KITTI & 61.2 & 72.6 \\
            KITTI & Ours & 39.2 & 48.3 \\
            \bottomrule
        \end{tabular}
	\end{minipage}%
	\begin{minipage}[t]{.05\linewidth}
		\ 
	\end{minipage} 
	\begin{minipage}[t]{.2\linewidth}
		\centering
		\caption{Results for lane instance segmentation}
		\label{tab:results_instance_segmentation}
		\begin{tabular}{lrr}
			\toprule
			Metric & Score\\
			\midrule
			AP & 0.250 \\
			AP@50 & 0.507 \\
			\bottomrule
		\end{tabular}
	\end{minipage} 
	
\end{table}

\subsection{Lane Instance Segmentation}
The annotation of multiple distinct lanes per image, the number of which is variable across images and potentially sequences, naturally suggests an instance segmentation task for our dataset. Though it has been postulated that ``Stuff'' is uncountable and therefore doesn't have instances~\cite{cocostuff,seeingstuff2010}, we present this lane instance segmentation task as a counter example. Indeed it would seem many stuff-like classes (parking spaces, lanes in a swimming pool,  fields in satellite imagery) can have meaningful delineations and therefore instances applied.

Providing a useful baseline for this lane instance segmentation task presents its own challenges. The current state of the art for instance segmentation on Cityscapes is MaskRCNN~\cite{DBLP:journals/corr/HeGDG17}. This approach is based on the RCNN object detector and is therefore optimised for the detection of compact objects which fit inside broadly non overlapping bounding boxes, traditionally called ``Things". In the case of lanes detected in the perspective view, a bounding box for any given lane greatly overlaps neighbouring lanes, making the task potentially challenging for standard bounding boxes. This becomes more apparent when the road undergoes even a slight curve in which case the bounding boxes are almost on top of one another even though the instance pixels are quite disjoint. Recently, a few works have explored an alternative approach to RCNN based algorithms which use pixel embeddings to perform instance segmentation~\cite{embeddingDiscLoss2017,1801.00908,DBLP:journals/corr/FathiWRWSGM17,1712.08273}; we provide a baseline for our dataset using pixel embeddings.

Specifically we train a model based on~\cite{embeddingDiscLoss2017}. We follow their approach of learning per pixel embeddings whose value is optimised such that pixels within the same training instance are given similar embeddings, while the mean embedding of separate instances are simultaneously pushed apart. A cost function which learns such pixel embeddings can be written down exactly and is presented in Eq.~1-4 of~\cite{embeddingDiscLoss2017}, we use the same hyper parameters reported in that work, and thus use an 8-dimensional embedding space.  We impose this loss as an extra output of a ROAD SegNet model trained along side the segmentation task from scratch. 

At run time we follow a variant of the approach proposed by~\cite{embeddingDiscLoss2017}, predicting an embedding per pixel. We use our prediction of road to filter away pixels which are not likely to be lanes. We then uniformly sample pixels in the road area and cluster their embeddings using the Mean Shift~\cite{Comaniciu:2002:MSR:513073.513076} algorithm, identifying the centres of our detected lane instances. Finally, all pixels in the road area are assigned to their closest lane instance embedding using the euclidean distance to the pixel's own embedding; pixels assigned to the same centroid are in the same instance. 

\begin{figure}
	\centering
	\includegraphics[width=0.9\linewidth]{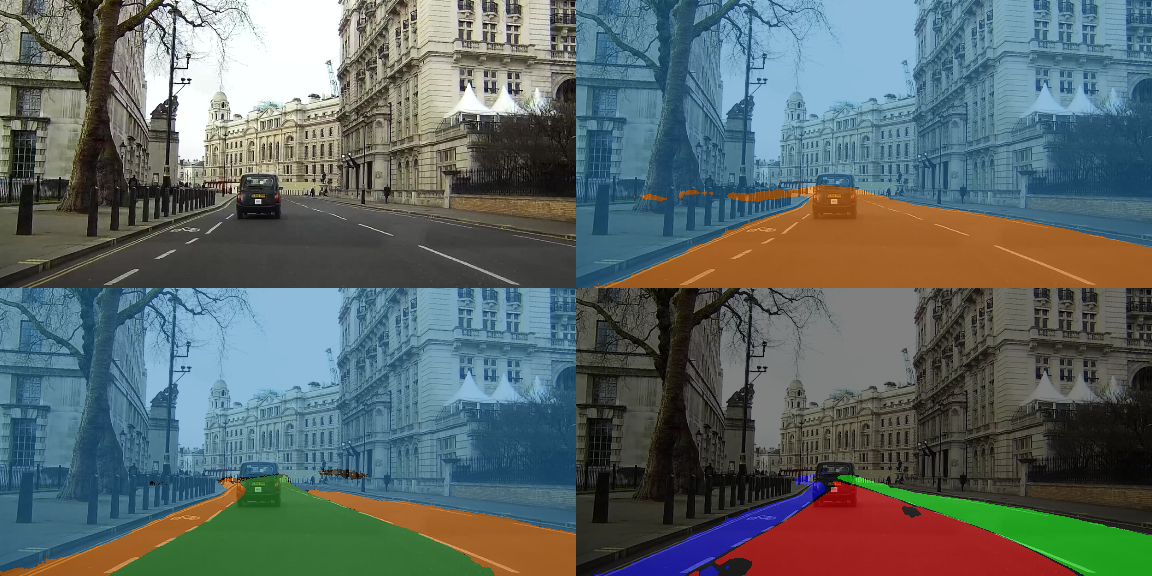}
	\caption{An example image from our test set (top left) including predictions for the ROAD (top right), EGO (bottom left) and instance (bottom right) tasks. The colours of the ROAD and EGO models match those in Figure~\ref{fig:streetscene}. The predicted instances are represented by red, green and blue.}
	\label{fig:londoncab}
\end{figure}

For evaluation, we use the Average Precision (AP) measures calculated as described for the MS-COCO \cite{lin2014microsoft} instance segmentation task. Specifically: we calculate the AP across images and across IoU thresholds of detected lanes (pixels assigned to embedding cluster centroids) and ground truth lanes. True and false positives are counted in the following way: (1) A detection is a \textbf{true positive} when it overlaps a ground truth instance with an IoU above some threshold and (2) a detection is a \textbf{false positive} when it does not sufficiently overlap any ground truth instance. Using these definitions we report average precision at 50\% IoU and an average AP across multiple thresholds from 50\% to 95\% in increments of 5\%.
Tab.~\ref{tab:results_instance_segmentation} shows the instance segmentation baseline results. Qualitatively, the lane instances are well separated, as can be seen in Fig.~\ref{fig:londoncab}.
%As our proposed approach does not provide a score for any given lane instance detection we instead order lane detections by their size, choosing to assign larger lane instances to ground truths before smaller ones.

\section{Conclusions}

We have created a dataset for road detection and lane instance segmentation in urban environments, using only un-calibrated low-cost equipment.
Moreover, we have done this using an efficient annotation procedure that minimises manual work.
The initial experiments presented show promising generalisation results across datasets.
Despite this step towards autonomous driving systems, our data has various limitations:
(1) Annotations of many other object classes of the static road layout are not included, like buildings, traffic signs and traffic lights.
(2) All annotated lanes are parallel to the future driven path, thus currently lane splits and perpendicular lanes (e.g. at junctions) have been excluded.
(3) Positions of dynamic objects, like vehicles, pedestrians and cyclists, are not included.
In future work, those limitations could be addressed by adding further annotations of different objects in 3D, inspired by \cite{xie2016semantic}.
Non-parallel lanes could be handled by extending our annotator tool to allow for variable angles for the lanes in the road plane.
Also, a pre-trained segmentation model could be used to better initialise the annotations.
Furthermore, the position of dynamic objects could be  estimated by including additional sensor modalities, like stereo vision or LIDAR.

\section*{Acknowledgements}
We would like to thank our colleagues Tom Westmacott, Joel Jakubovic and Robert Chandler, who have contributed to the implementation of the annotation software.

\clearpage

\bibliographystyle{splncs}
\bibliography{library,library_manual}
\end{document}